# Intrusion Detection using Continuous Time Bayesian Networks

**Jing Xu**                                                    JINGXU@CS.UCR.EDU

**Christian R. Shelton**                                    CSHELTON@CS.UCR.EDU

*Department of Computer Science and Engineering*

*University of California, Riverside*

*Riverside, CA 92521, USA*

## Abstract

Intrusion detection systems (IDSs) fall into two high-level categories: network-based systems (NIDS) that monitor network behaviors, and host-based systems (HIDS) that monitor system calls. In this work, we present a general technique for both systems. We use anomaly detection, which identifies patterns not conforming to a historic norm. In both types of systems, the rates of change vary dramatically over time (due to burstiness) and over components (due to service difference). To efficiently model such systems, we use continuous time Bayesian networks (CTBNs) and avoid specifying a fixed update interval common to discrete-time models. We build generative models from the normal training data, and abnormal behaviors are flagged based on their likelihood under this norm. For NIDS, we construct a hierarchical CTBN model for the network packet traces and use Rao-Blackwellized particle filtering to learn the parameters. We illustrate the power of our method through experiments on detecting real worms and identifying hosts on two publicly available network traces, the MAWI dataset and the LBNL dataset. For HIDS, we develop a novel learning method to deal with the finite resolution of system log file time stamps, without losing the benefits of our continuous time model. We demonstrate the method by detecting intrusions in the DARPA 1998 BSM dataset.

## 1. Introduction

Misuse or abuse of computer systems is a critical issue for system administrators. Our goal is to detect these attacks that attempt to compromise the performance quality of a particular host machine.

It is time-consuming and error-prone to acquire labeled data that contains both good and bad behaviors from which to build a classifier. Additionally, the frequency with which attacks are developed can make maintaining a database of all previously seen attacks inefficient or even infeasible. Anomaly detection can identify new attacks even if the attack type was unknown beforehand. Unsupervised learning allows the anomaly detector to adapt to changing environments, thereby extending its domain of usefulness. By modeling normal behavior from historic clean data, we can identify abnormal activity without a direct prior model of the attack by simply comparing its deviation from the learned norm.

In a network-based intrusion detection system (NIDS), the network packet traces are monitored. Network traffic traces collect information from a network's data stream and provide an external view of the network behavior. In a host-based intrusion detection system (HIDS), the internal state of a computing system is analyzed. System call logs are a convenient way of monitoring executing programs' behavior through their operating system calls.

Both systems are composed of activities that happen at dramatically different time granularity. Users alternate between busily using their computer and resting. During the busy period, a burst of action may cause a peak of network traffic flow or operating system usage. However, during the resting period, the computer just maintains its regular running pattern, and network or system activities are much less intense, *e.g.* automatically checking email every few minutes. Even within each of these global modes there are variations. Therefore, a dynamic model that requires discretizing the time is not efficient. We develop intrusion detection techniques using continuous time Bayesian networks (CTBNs) (Nodelman, Shelton, & Koller, 2002) for both data types. Although the two data are of completely different formats and semantic meaning, we demonstrate the flexibility of a continuous time generative model (such as a CTBN) to describe either.

Our first effort is to detect anomalies from network traffic traces (NIDS). Abnormal traffic must differ in some way from the normal traffic patterns. While this difference may be very subtle and difficult to detect, the more subtle the attack, the longer the attack will take and the more it will stress the patience of the attacker. Looking at summarized information like flow statistics is not helpful, especially for stealthy worms which can mingle well with normal traffic by sacrificing their spreading speed and scale. We, therefore, feel that looking for abnormalities in the detailed network traffic flow level is a utile method for finding attacks. A network flow for a given host machine is a sequence of continuous-time asynchronous events. Furthermore, these events form a complex structured system, where statistical dependencies relate network activities like packet emissions and connections. We employ CTBNs to reason about these structured stochastic network processes.

Our CTBN model contains a number of observed network events (packet emissions and concurrent port connections changes). To allow our model to be more descriptive, we also add latent variables that tie the activity variables together. Exact inference in this method is no longer feasible. Therefore, we use Rao-Blackwellized particle filtering (RBPF) to estimate the parameters.

Our second effort is to detect intrusions using system call logs (HIDS). A system log file contains an ordered list of calls made to a computer's operating system by an executing program. We focus on analyzing the ordering and the context of the sequence, rather than simply counting the overall statistics. A CTBN is a natural way of modeling such sequential data. Because of the finite resolution of computer clock, all the system calls issued within a clock tick are assigned a same time stamp. Therefore the data stream consists of long periods of time with no activity, followed by sequences of calls in which the order is correctly recorded, but the exact timing information is lost. This poses a new challenge for CTBN reasoning. We present here a learning method for such type of data without resorting to time discretization.

We validate our NIDS technique on the MAWI dataset and the LBNL dataset, and our HIDS technique on the DARPA 1998 BSM dataset. Both applications give good results when compared with other method.

In Section 2 we discuss the related work in intrusion detection. In Section 3 we review continuous-time Markov processes and continuous time Bayesian networks. In Section 4 we describe the CTBN model and the RBPF inference algorithms for the NIDS problem. In Section 5 we describe the CTBN model and the parameter estimation algorithm for HIDS, including how to deal with imprecise timing measurements. In Section 6 we show our experimental results for both of the applications.

## 2. Related Work

Much of the previous work in intrusion detection focuses on one area only — either detecting the network traffic or mining the system call logs. The work of Eskin, Arnold, Prerau, Portnoy, and Stolfo (2002) is similar to our approach in that they apply their method to both of these kinds of data. They map data elements to a feature space and detect anomalies by determining which points lie in sparse regions using cluster-based estimation, K-nearest neighbors and one-class SVM. They use a data-dependent normalization feature map for network traffic data and a spectrum kernel for system call traces.

## 2.1 NIDS

For network traffic data, we build upon our previous work (Xu & Shelton, 2008). There we made the assumption that network activities are independent across different ports. This allowed us to factorize the model into port-level submodels and standard exact inference techniques could be used for parameter learning. In this paper, we remove this restriction. There is no application-specific reason that traffic should be independent by ports. By tying the traffic together, our model describes more complicated structural dependencies among variables. We derive a Rao-Blackwellized particle filtering algorithm to estimate the parameters for our model. Our work also differs in that we are not only interested in the intrusion detection problem, but host identity recognition as well.

As a signature-based detection algorithm, we share many of the assumptions of Karagiannis, Papagiannaki, and Faloutsos (2005). In particular, we also assume that we do not have access to the internals of the machines on the networks, which rules out methods like those of Malan and Smith (2005), Cha (2005), Qin and Lee (2004), and Eskin et al. (2002). However, we differ in that our approach does not rely on preset values, require human intervention and interpretation, nor assume that we have access to network-wide traffic information. Network-wide data and human intervention have advantages, but they can also lead to difficulties (data collation in the face of an attack and increased human effort), so we chose to leave them out of our solution.

Many learning, or adaptive, methods have been proposed for network data. Some of these — for example, those of Zuev and Moore (2005) and Soule, Salamatian, Taft, Emilion, and Papagiannali (2004) — approach the problem as a classification task which requires labeled data. Dewaele, Fukuda, and Borgnat (2007) profile the statistical characteristics of anomalies by using random projection techniques (sketches) to reduce the data dimensionality and a multi-resolution non-Gaussian marginal distribution to extract anomalies at different aggregation levels. The goal of such papers is usually not to detect attacks but rather to classify non-attacks by traffic type; if applied to attack detection, they would risk missing new types of attacks. Furthermore, they frequently treat each network activity separately, instead of considering their temporal context.

Lakhina, Crovella, and Diot (2005) has a nice summary of adaptive (or statistical) methods that look at anomaly detection (instead of classification). They use an entropy-based method for the entire network traffic. Many of the other methods, such as that of Ye, Emran, Chen, and Vilbert (2002), use either statistical tests or subspace methods that assume the features of the connections or packets are distributed normally. Rieck and Laskov (2007) model the language features like n-grams and words from connection payloads. Xu, Zhang, and Bhattacharyya (2005) also use unsupervised methods, but they concentrate on clustering traffic across a whole network. Similarly, Soule, Salamatian, and Taft (2005) build an anomaly detector based on Markov models, but it is for the network traffic patterns as a whole and does not function at the host level.

The work of Soule et al. (2004) is very similar in statistical flavor to our work. They also fit a distribution (in their case, a histogram modeled as a Dirichlet distribution) to network data. However, they model flow-level statistics, whereas we work at the level of individual connections. Additionally, they are attempting network-wide clustering of flows instead of anomaly detection.

The work of Moore and Zuev (2005), like our approach, models traffic with graphical models, in particular, Naive Bayes networks. But their goal is to categorize network traffic instead of detecting attacks. Kruegel, Mutz, Robertson, and Valeur (2003) present a Bayesian approach to the detecting problem as an event classification task while we only care about whether the host is under attack during an interval.

The work of Lazarevic, Ertoz, Kumar, Ozgur, and Srivastava (2003) is also similar to our work. It is one of the few papers to attempt to find attacks at the host level. They employ nearest neighbor, a Mahalanobis distance approach, and a density-based local outliers method, each using 23 features of the connections. Although their methods make the standard *i.i.d.* assumption about the data (and therefore miss the temporal context of the connection) and use 23 features (compared to our few features), we compare our results to theirs in Section 6, as the closest prior work. Agosta, Duik-Wasser, Chandrashekar, and Livadas (2007) present an adaptive detector whose threshold is time-varying. It is similar to our work in that they also rely on model-based algorithms. But they employ the host internal states like CPU loads which are not available to us.

While there has been a great variety of previous work, our work is novel in that it detects anomalies at the host level using only the timing features of network activities. We do not consider each connection (or packet) in isolation, but rather in a complex context. We capture the statistical dynamic dependencies between packets and connections to find sequences of network traffic that are anomalous *as a group*.

## 2.2 HIDS

Previous work on detecting intrusions in system call logs can be roughly grouped into two categories: sequence-based and feature-based. Sequence-based methods focus on the sequential order of the events while feature-based methods treat system calls as independent data elements. Our method belongs to the former category since we use a CTBN to model the dynamics of the sequences.

Time-delay embedding (tide) and sequence time-delay embedding (stide) are two examples of sequence based methods (Forrest, A.Hofmeyr, Somayaji, & A.Longstaff, 1996; A.Hofmeyr, Forrest, & Somayaji, 1998). They generalize the data by building a database storing previously seen system call sub-sequences, and test by looking up subsequences in the database. These methods are straightforward and often achieve good results. We compare with them in our experiments. Tandon and Chan (2005) look at a richer set of attributes like return value and arguments associated with a system call while we only make use of the system call names.

Feature based methods like those of Hu, Liao, and Vemuri (2003) use the same dataset we use, the DARPA 1998 BSM dataset, but their training data is noisy and they try to find a classification hyperplane using robust support vector machines (RSVMs) to separate normal system call profiles from intrusive ones. Eskin (2000) also works on noisy data. They make the assumption that their training data contains a large portion of normal elements and few anomalies. They present a mixture of distribution over normal and abnormal data and calculate the likelihood change if a data point is moved from normal part to abnormal part to get the optimum data partition.

Yeung and Ding (2002) try to use both techniques. They provide both dynamic and static behavioral models for system call data. For the dynamic method, a hidden Markov model (HMM) is used to model the normal system events and a likelihood is calculated for each testing sequence and compared against a certain threshold. Our work for the system call traces problem is very close to their framework since we also build a dynamic model for the sequential data and compute the likelihood of a testing example as a score. But we are different in that our CTBN models the continuous time dynamics rather than time-sliced behaviors. For the static method, they represent the normal behavior by a command occurrence frequency distribution and measure the distance from the testing example to this norm by cross entropy. The dataset they use is KDD archive dataset.

## 2.3 Other Work

Simma et al. (2008) also use a continuous-time model to reason about network traffic. They apply their method to find dependences in exterprise-level services. Their model is non-Markovian, but also deals with network events as the basic observational unit.

To estimate the parameters of the large network we build for the network traffic data, we use Rao-Blackwellized particle filters (RBPFs). Doucet, de Freitas, Murphy, and Russel (2000) propose a RBPF algorithm for dynamic Bayesian networks that works in discrete time fashion by exploiting the structure of the DBN. Ng, Pfeffer, and Dearden (2005) extend the RBPF to continuous time dynamic systems and apply the method to the K-9 experimental Mars rover at NASA Ames Research Center. Their model is a hybrid system containing both discrete and continuous variables. They use particle filters for the discrete variables and unscented filters for the continuous variables. Our work are similar to theirs in that we apply a RBPF to a CTBN. But our model only contains discrete variables and our evidence is over continuous time (as opposed to only "snapshots" of the system state).

# 3. Continuous Time Bayesian Networks

We begin by briefly reviewing the definition of Markov processes and continuous time Bayesian networks (CTBNs).

## 3.1 Homogeneous Markov Process

A finite-state, continuous-time, homogeneous Markov process $X_t$ is described by an initial distribution $P_X^0$ and, given a state space $Val(X) = \{x_1, ..., x_n\}$, an $n \times n$ matrix of transition intensities:

$$\mathbf{Q_X} = \begin{bmatrix} -q_{x_1} & q_{x_1 x_2} & \cdots & q_{x_1 x_n} \\ q_{x_2 x_1} & -q_{x_2} & \cdots & q_{x_2 x_n} \\ \vdots & \vdots & \ddots & \vdots \\ q_{x_n x_1} & q_{x_n x_2} & \cdots & -q_{x_n} \end{bmatrix}.$$

$q_{x_i x_j}$ is the intensity (or rate) of transition from state $x_i$ to state $x_j$ and $q_{x_i} = \sum_{j \neq i} q_{x_i x_j}$.

The transient behavior of $X_t$ can be described as follows. Variable $X$ stays in state $x$ for time exponentially distributed with parameter $q_x$. The probability density function $f$ for $X_t$ remaining at $x$ for duration $t$ is $f_x(q, t) = q_x \exp(-q_x t)$ for $t \geq 0$. The expected time to the next transition given the state is currently $x$ is $1/q_x$. Upon transitioning, $X$ shifts to state $x'$ with probability $\theta_{xx'} = q_{xx'}/q_x$. Note that given $q_x$, $\theta_{xx'}$ and $q_{xx'}$ are iosmorphic. We will sometime gives formulae in terms of $\theta_{xx'}$ where it simplifies the expression.

The distribution over the state of the process $X$ at some future time $t$, $P_x(t)$, can be computed directly from $\mathbf{Q_X}$. If $P_X^0$ is the distribution over $X$ at time 0 (represented as a vector), then, letting exp be the matrix exponential,

$$P_X(t) = P_X^0 \exp(\mathbf{Q_X} \cdot \mathbf{t}) \ .$$

## 3.2 Complete Data

Complete data for an HMP are represented by a set of trajectories $\mathbf{D} = \{\tau_1, ... \tau_n\}$. Each trajectory $\tau_i$ is a complete set of state transitions: $d = \{(x_d, t_d, x'_d)\}$, meaning that $X$ stayed in state $x_d$ for a duration of $t_d$, and then transitioned to state $x'_d$. Therefore we know the exact state of the variable $X$ at any time $0 \le t \le T$.

## 3.3 Sufficient Statistics and Likelihood

Given an HMP and its full data $\mathbf{D}$, the likelihood of a single state transition $d = \{(x_d, t_d, x'_d)\} \in \mathbf{D}$ is

$$L_X(q, \theta : d) = (q_{x_d} \exp(-q_{x_d} t_d))(\theta_{x_d x'_d}) \ .$$

The likelihood function for $\mathbf{D}$ can be decomposed by transition:

$$\begin{aligned}
L_X(q, \theta : \mathbf{D}) &= (\prod_{d \in \mathbf{D}} L_X(q : d))(\prod_{d \in \mathbf{D}} L_X(\theta : d)) \\
&= (\prod_x q_x^{M[x]} \exp(-q_x T[x]))(\prod_x \prod_{x' \ne x} \theta_{xx'}^{M[x,x']}) \ .
\end{aligned}$$

If we take the log of the above function, we get the log likelihood:

$$\begin{aligned}
l_X(q, \theta : \mathbf{D}) &= l_X(q : \mathbf{D}) + l_X(\theta : \mathbf{D}) \\
&= \sum_x (M[x] \ln(q_x) - q_x T[x] + \sum_{x' \ne x} M[x, x'] \ln(\theta_{xx'})) \ .
\end{aligned}$$

Here $M[x, x']$ and $T[x]$ are the *sufficient statistics* of the HMP model. $M[x, x']$ is the number of times $X$ transitions from the state $x$ to $x'$. We denote $M[x] = \sum_{x'} M[x, x']$, the total number of times the system leaves state $x$. $T[x]$ is the total duration that $X$ stays in the state $x$.

## 3.4 Learning from Complete Data

To estimate the parameters of the transition intensity matrix $Q$, we maximize the above log likelihood function. This yields the maximum likelihood estimates:

$$\hat{q}_x = \frac{M[x]}{T[x]}, \quad \hat{\theta}_{xx'} = \frac{M[x, x']}{M[x]} \ .$$

## 3.5 Incomplete Data

Incomplete data from an HMP are composed partially observed trajectories $\mathbf{D} = \{\tau_1^-, ... \tau_n^-\}$. Each trajectory $\tau_i^-$ consists of a set of $d = \{(S_d, t_d, dt)\}$ observations, where $S_d$ is a *subsystem* (a nonempty subset of the states of $X$) of the process. Each of the triplets specifies an "interval evidence." It states that the variable $X$ is in the subsystem $S_d$ from time $t_d$ to time $t_d + dt$. Some of the observations may be duration-free. *i.e.*, we only observe $X \in S_d$ at time $t$, but do not know how long it stayed there. This is called a "point evidence" and can be generalized using the same triplet notation described above by setting the duration to be 0. For a partially observed trajectory, we only observe sequences of subsystems, and do not observe the state transitions within the subsystems.

### 3.6 Expected Sufficient Statistics and Expected Likelihood

We can consider possible completions of a partially observed trajectory that specify the transitions that are consistent with the partial trajectory. By combining the partial trajectory and its completion, we get a full trajectory. We define $\mathbf{D}^+ = \{\tau_1^+, ..., \tau_n^+\}$ to be completions of all the partial trajectories in $\mathbf{D}$. Given a model, we have a distribution over $\mathbf{D}^+$, given $\mathbf{D}$.

For data $\mathbf{D}^+$, the expected sufficient statistics with respect to the probability density over possible completions of the data are $\bar{T}[x]$, $\bar{M}[x, x']$ and $\bar{M}[x]$. The expected log likelihood is

$$E[l_X(q, \theta : \mathbf{D}^+)] = E[l_X(q : \mathbf{D}^+)] + E[l_X(\theta : \mathbf{D}^+)]$$
$$= \sum_x (\bar{M}[x] \ln(q_x) - q_x \bar{T}[x] + \sum_{x' \neq x} \bar{M}[x, x'] \ln(\theta_{xx'})) .$$

### 3.7 Learning from Incomplete Data

The expectation maximization (EM) algorithm can be used to find a local maximum of the likelihood from partial trajectory. The EM algorithm iterates over the following E step and M step until the convergence on the derived likelihood function.

E step: Given the current HMP parameters, compute the expected sufficient statistics: $\bar{T}[x]$, $\bar{M}[x, x']$ and $\bar{M}[x]$ for the data set $\mathbf{D}$. This is the most complex part of the algorithm. We give further details below.

M step: From the computed expected sufficient statistics, update the new model parameters for the next EM iteration:

$$q_x = \frac{\bar{M}[x]}{\bar{T}[x]}, \quad \theta_{xx'} = \frac{\bar{M}[x, x']}{\bar{M}[x]} .$$

Now we show how to calculate the expected sufficient statistics using the forward-backward message passing method.

A trajectory $\tau \in \mathbf{D}$ can be devided into $N$ intervals where each of the interval is separated by adjacent event changes. Assume the trajectory spans the time interval $[0, T)$, and let $\tau[v, w]$ be the observed evidence between time $v$ and $w$, including events on the time stamp $v$ and $w$, and let $\tau(v, w)$ be the same set of evidence but excluding $v$ and $w$. Let $S$ be the subsystem the states are restricted on this interval.

We define

$$\alpha_{\mathbf{t}} = P(X_t, \tau[0, t]), \quad \beta_{\mathbf{t}} = P(\tau[t, T] \mid X_t)$$

to be vectors (indexed by possible assignments to $X_t$). Similarly, we define the corresponding distribution that excludes certain point evidence as follows.

$$\alpha_{\mathbf{t}}^- = P(X_t, \tau[0, t)), \quad \beta_{\mathbf{t}}^+ = P(\tau(t, T] \mid X_t) .$$

Denote $\delta_j$ to be a vector of all 0's except for its $j$-th position being 1, and denote $\mathbf{\Delta_{ij}}$ be a matrix of all 0's except that the element on $i$-th row and $j$-th column is 1.

We are now able to show the derived expected sufficient statistics. For time,

$$E[T[x]] = \int_0^T P(X_t \mid \tau[0, T]) \delta_x \, dt$$
$$= \frac{1}{P(\tau[0, T])} \sum_{i=0}^{N-1} \int_{t_i}^{t_{i+1}} P(X_t, \tau[0, T]) \delta_x \, dt .$$

The constant fraction at the beginning of the last line serves to make the total expected time over all j sum to $\tau$. The integral on each interval can be further expressed as

$$\int_v^w P(X_t, \tau[0,T])\delta_x \, dt = \int_v^w \alpha_{\mathbf{v}} \exp(\mathbf{Q_S}(t-v)) \Delta_{\mathbf{xx}} \exp(\mathbf{Q_S}(w-t))\beta_{\mathbf{w}} \, dt \ ,$$

where $Q_S$ is the same as $Q_X$ except all elements that correspond to transitions to or from $S$ are set to 0.

The equation for expected transition counts can similarly be defined:

$$E[M[x,x']] = \frac{\mathbf{q_{x,x'}}}{P(\tau[0,T])} \big[ \sum_{i=1}^{N-1} \alpha^-_{\mathbf{t_i}} \Delta_{\mathbf{x,x'}} \beta^+_{\mathbf{t+i}}$$
$$+ \sum_{i=0}^{N-1} \int_{t_i}^{t_{i+1}} \alpha_{\mathbf{t_i}} \exp(\mathbf{Q_S}(t-t_i)) \Delta_{\mathbf{x,x'}} \exp(\mathbf{Q_S}(t_{i+1}-t))\beta_{\mathbf{t_{i+1}}} \, dt \big] \ .$$

The integrals appearing in $E[T]$ and $E[M]$ can be computed via a standard ODE solver, like the Runge-Kutta method (Press, Teukolsky, Vetterling, & Flannery, 1992). Such a method uses an adaptive step size to move quickly through times of few expected changes and more slowly through times of rapid transitions.

Now the only remaining problem is to calculate $\alpha$ and $\beta$. Let $\mathbf{Q_{SS'}}$ be the transitioning intensity matrix of the HMP from one subsystem $S$ to another $S'$. This matrix is the same as $Q_X$, but only elements corresponding to transitions from $S$ to $S'$ are non-zero.

$$\alpha^-_{\mathbf{t_i}} = \alpha_{\mathbf{t_{i-1}}} \exp(\mathbf{Q_{S_{i-1}}}(t_i - t_{i-1})) \ ,$$
$$\alpha_{\mathbf{t_i}} = \alpha^-_{\mathbf{t_i}} \mathbf{Q_{S_{i-1}S_i}} \ ,$$
$$\beta^-_{\mathbf{t_i}} = \exp(\mathbf{Q_{S_i}}(t_{i+1} - t_i))\beta_{\mathbf{t_{i+1}}} \ ,$$
$$\beta_{\mathbf{t_i}} = \mathbf{Q_{S_{i-1}S_i}}\beta^-_{\mathbf{t_i}} \ .$$

During this forward-backward calculation, it is also trivial to answer queries such as

$$P(X_t = x \mid \tau[0,T]) = \frac{1}{P(\tau)} \alpha^-_{\mathbf{t}} \Delta_{\mathbf{xx}}\beta_{\mathbf{t}} \ .$$

## 3.8 Continuous Time Bayesian Networks

While HMPs are good for modeling many dynamic systems, they have their limitations when the systems have multiple components because the state space grows exponentially in the number of variables. An HMP does not model the variable independencies and therefore it has to use a unified state $X$ to represent the joint behavior of all the involving components in the system. In the this section, we show how a continuous time Bayesian network can be used to address this issue.

Nodelman et al. (2002) extend the theory of HMPs and present continuous time Bayesian networks (CTBNs), which model the joint dynamics of several local variables by allowing the transition model of each local variable $X$ to be a Markov process whose parametrization depends on some subset of other variables $U$.

### 3.9 Definition

We first give an definition of an inhomogeneous Markov process called a conditional Markov process. It is a critical concept for us to formally introduce the CTBN framework.

**Definition 1** (Nodelman, Shelton, & Koller, 2003) *A conditional Markov process $X$ is an inhomogeneous Markov process whose intensity matrix varies as a function of the current values of a set of discrete conditioning variables $U$. It is parametrized using a conditional intensity matrix (CIM) $Q_{X|U}$ – a set of homogeneous intensity matrices $Q_{X|u}$, one for each instantiation of values $u$ to $U$.*

We call $U$ the *parents* of $X$. When the set of $U$ is empty, the CIM is simply a standard intensity matrix.

CIMs provide a way to model the temporal behavior of one variable conditioned on some other variables. By putting these local models together, we have a joint structured model — a continuous time Bayesian network.

**Definition 2** (Nodelman et al., 2003) *A continuous time Bayesian network $\mathcal{N}$ over a set of stochastic processes $\boldsymbol{X}$ consists of two components: an initial distribution $P_{\boldsymbol{X}}^0$, specified as a Bayesian network $\mathcal{B}$ over a set of random variables $\boldsymbol{X}$, and a continuous transition model, specified using a directed (possibly cyclic) graph $\mathcal{G}$ whose nodes are $X \in \boldsymbol{X}$; $\boldsymbol{U}_X$ denotes the parents of $X$ in $\mathcal{G}$. Each variable $X \in \boldsymbol{X}$ is associated with a conditional intensity matrix, $\boldsymbol{Q}_{X|\boldsymbol{U}_X}$.*

The dynamics of a CTBN are quantitatively defined by a graph. The instantaneous evolution of a variable depends only on the current value of its parents in the graph. The quantitative description of a variable's dynamics is given by a set of intensity matrices, one for each value of its parents. That means the transition behavior of the variable is controlled by the current values of its parents.

The standard notion of $d$-separation from Bayesian networks carries over to CTBNs. Because graphs are cyclic and variables represent processes (not single random variables), the implications are a little different. A variable (process) is still independent of its non-descendants given its parents, and it is still independent of everything given its Markov blanket (any variable that is either a parent, a child, or a parent of a child). Cycles can cause parents to also be children, but provided they are considered as both, the above definitions still hold. More importantly, the notion of "given" works only if the *full* trajectory for the variable in question is known. Therefore, $X$ and its grandchildren are *not* independent given $X$'s children's values at a single instant. Rather, they are only independent given $X$'s children's full trajectories from time 0 until the last time of interest.

If we amalgamate all the variables in the CTBN together, we get a single homogeneous Markov process over the joint state space. In the joint state intensity matrix, a rate of 0 is assigned to any transition that involves changing more than one variable's value at the exact same time. All other intensities can be found by looking up the value in the corresponding conditional intensity matrix for the variable that changes. The diagonal elements are the negative row sums.

Forward sampling can be done quickly in a CTBN without generating the full joint intensity matrix. We keep track of the "next event time" for each variable (sampled from the relevant exponential distribution given the current values of itself and its parent). We then select the earliest event time and change that variable (sampling from the multinomial distribution implied by the row of that variable's relevant intensity matrix). The next event time for the variable that just changed and all of its children must be resampled, but no other variable's time must be resampled due to the memoryless property of the exponential distribution. In this way a sequence of events (a trajectory) can be sampled.

### 3.10 Learning

In the context of CTBNs, the model parameters consist of the CTBN structure $\mathcal{G}$, the initial distribution $P_0$ parameterized by a regular Bayesian network, and the conditional intensity matrices (CIMs) of each variable in the network. In this section, we assume the CTBN structure is known to us, so we only focus on the parameter learning. We also assume the model is irreducible. So the initial distribution $P_0$ becomes less important in the context of CTBN inference and learning, especially when the time range becomes significantly large. Therefore, parameter learning in our context is to estimate the conditional intensity matrices $Q_{X_i|\mathbf{U_i}}$ for each variable $X_i$, where $\mathbf{U_i}$ is the set of parent variables of $X_i$.

#### 3.10.1 Learning from Complete Data

Nodelman et al. (2003) presented an efficient way to learn a CTBN model from fully observed trajectories. With complete data, we know full instantiations to all the variables for the whole trajectory. So we know which CIM is governing the transition dynamics of each variable at any time. The sufficient statistics are $M[x, x'|u]$ — the number of times $X$ transitions from the state $x$ to $x'$ given its parent instantiation $u$ — and $T[x|u]$ — the total duration that $X$ stays in the state $x$ given its parent instantiation $u$. We denote $M[x|u] = \sum_{x'} M[x, x'|u]$.

The likelihood function for $\mathbf{D}$ can be decomposed as

$$L_{\mathcal{N}}(q, \theta : \mathbf{D}) = \prod_{X_i \in \mathbf{X}} L_{X_i}(q_{X_i|U_i} : \mathbf{D}) L_{X_i}(\theta_{X_i|U_i} : \mathbf{D})) \tag{1}$$

where

$$L_X(q_{X|\mathbf{U}} : \mathbf{D}) = \prod_{\mathbf{u}} \prod_{x} q_{x|\mathbf{u}}^{M[x|\mathbf{u}]} \exp(-q_{x|\mathbf{u}} T[x|\mathbf{u}]) \tag{2}$$

and

$$L_X(\theta : \mathbf{D}) = \prod_{\mathbf{u}} \prod_{x} \prod_{x' \neq x} \theta_{xx'|\mathbf{u}}^{M}[x, x'|\mathbf{u}] \ . \tag{3}$$

If we put the above functions together and take the log, we get the log likelihood component for a single variable $X$:

$$
\begin{aligned}
l_X(q, \theta : \mathbf{D}) &= l_X(q : \mathbf{D}) + l_X(\theta : \mathbf{D}) \\
&= \sum_{\mathbf{u}} \sum_{x} M[x|u] \ln(q_x|u) - q_{[x|u]} T[x|u] \\
&+ \sum_{\mathbf{u}} \sum_{x} \sum_{x' \neq x} M[x, x'|u] \ln(\theta_{xx'|u})).
\end{aligned} \tag{4}
$$

By maximizing the above log likelihood function, the model parameters can be estimated as

$$\hat{q}_{x|\mathbf{u}} = \frac{M[x|\mathbf{u}]}{T[x|\mathbf{u}]}, \quad \hat{\theta}_{xx'|\mathbf{u}} = \frac{M[x, x'|\mathbf{u}]}{M[x|\mathbf{u}]} \ . \tag{5}$$

### 3.10.2 LEARNING FROM INCOMPLETE DATA

Nodelman, Shelton, and Koller (2005) present the expectation maximization (EM) algorithm to learn a CTBN model from partially observed trajectories $\mathbf{D}$. The expected sufficient statistics are $\bar{M}[x, x'|\mathbf{u}]$, the expected number of times that $X$ transitions from state $x$ to $x'$ when its parent set $\mathbf{U}$ takes the values $\mathbf{u}$, and $\bar{T}[x|\mathbf{u}]$, the expected amount of time that $X$ stays in the state $x$ under the parent instantiation $\mathbf{u}$. We denote $\bar{M}[x|\mathbf{u}]$ to be $\sum_{x'} \bar{M}[x, x'|u]$. The expected log likelihood can be decomposed in the same way as in Equation 4, except that the sufficient statistics $M[x, x'|u]$, $T[x|u]$ and $M[x|u]$ are now replaced with expected sufficient statistics $\bar{M}[x, x'|u]$, $\bar{T}[x|u]$ and $\bar{M}[x|u]$.

The EM algorithm for a CTBN works essentially in the same way as for an HMP. The expectation step is to calculate the expected sufficient statistics using inference method (will be described in Section 3.11). The maximization step is to update the model parameters:

$$\hat{q}_{x|\mathbf{u}} = \frac{\bar{M}[x|\mathbf{u}]}{\bar{T}[x|\mathbf{u}]}, \quad \hat{\theta}_{xx'|\mathbf{u}} = \frac{\bar{M}[x, x'|\mathbf{u}]}{\bar{M}[x|\mathbf{u}]} \ .$$

## 3.11 Inference

Now given a CTBN model and some (partially) observed data, we would like to query the model. For example, we may wish to calculate the expected sufficient statistics for the above EM algorithm.

### 3.11.1 EXACT INFERENCE

Nodelman et al. (2005) provide an exact inference algorithm using expectation maximization to reason and learn the parameters from partially observed data. This exact inference algorithm requires flattening all the variables into a single Markov process and performing inference as in an HMP. It has the problem that it makes the state space grow exponentially large. Therefore, the exact inference method is only feasible for problems with very small state spaces.

### 3.11.2 APPROXIMATE INFERENCE

Because of the issue addressed below, much work has been done on CTBN approximate inference. Nodelman, Koller, and Shelton (2005) present an expectation propagation algorithm. Saria, Nodelman, and Koller (2007) give another message passing algorithm that adapts the time granularity. Cohn, El-Hay, Friedman, and Kupferman (2009) provide a mean field variational approach. El-Hay, Friedman, and Kupferman (2008) show a Gibbs sampling method approach using Monte Carlo expectation maximization. Fan and Shelton (2008) give another sampling based approach that uses importance sampling. El-Hay, Cohn, Friedman, and Kupferman (2010) describe a different expectation propagation approach.

To estimate the parameters of the models we build for the two applications (NIDS and HIDS), we employ inference algorithms including exact inference and a Rao-Blackwellized particle filtering (RBPF) algorithm, depending on the model size. Ng et al. (2005) extended RBPF to CTBNs. Their model was a hybrid system containing both discrete and continuous variable. They used particle filters for the discrete variables and unscented filters for the continuous variable. Our work are similar to this work in the method of applying RBPF to CTBNs, but our contains only discrete variables and our evidence is over continuous intervals.

| PORT | DESCRIPTION | PORT | DESCRIPTION |
|------|-------------|------|-------------|
| 80 | World Wide Web HTTP | 80 | World Wide Wed HTTP |
| 8080 | HTTP Alternate | 139 | NETBIOS Session Service |
| 443 | HTTP protocol over TLS/SSL | 443 | HTTP protocol over TLS/SSL |
| 113 | Authentication Service | 445 | Microsoft-DS |
| 5101 | Talarian TCP | 1863 | MSNP |
| 995 | pop3 protocol over TLS/SSL | 2678 | Gadget Gate 2 Way |
| 51730 | *unknown* | 1170 | AT+C License Manager |
| 59822 | *unknown* | 110 | Post Office Protocol - Version 3 |

Figure 1: Ranking of the most frequent ports on MAWI dataset (left) and LBNL dataset (right).

## 3.12 CTBN Applications

Although inference and learning algorithms have been well developed for CTBNs, there have been only a few applications to real world problems. Nodelman and Horvitz (2003) used CTBNs to reason about users' presence and availability over time. Ng et al. (2005) used CTBNs to monitor a mobile robot. Nodelman et al. (2005) used CTBNs to model life event history. Fan and Shelton (2009) modeled social networks via CTBNs. Our previous work (Xu & Shelton, 2008) presented an NIDS for host machine using CTBNs, but did not include HIDS.

## 4. Anomaly Detection Using Network Traffic

In this section, we present an algorithm to detect anomalies in network traffic data using CTBNs. We only focus on a single host on the network. The sequence and timing of events (e.g. packet transimission and connection establishment) are very important in network traffic flow. It matters not just how many connections were initiated in the past minute, but also their timing: if they were evenly spaced the trace is probably normal, but if they all came in a quick burst it is more suspicious. Similarly, the sequence is important. If the connections were made to sequentially increasing ports it is more likely to be a scanning virus, whereas the same set of ports in random order is more likely to be normal traffic. These are merely simple examples. We would like to detect more complex patterns.

A typical machine in the network may have diverse activities with various service types (*e.g.* HTTP, SMTP). The destination port number roughly describes the type of service to which a particular network activity belongs. Some worms propagate malicious traffic toward certain well known ports to affect the quality of the associated services. By looking at traffic associated with different ports we are more sensitive to subtle variations that do not appear if we aggregate trace information across ports. Figure 1 shows the most popular ports ranked by their frequencies in the network traffic on the datasets we use (described in more depth later). These services are, to some extent, independent of each other. We therefore model each port's traffic with its own CTBN submodel. We denote $\tau$ as the whole observed traffic sequences on the particular host, and $\tau_j$ as the traffic associated with port $j$.

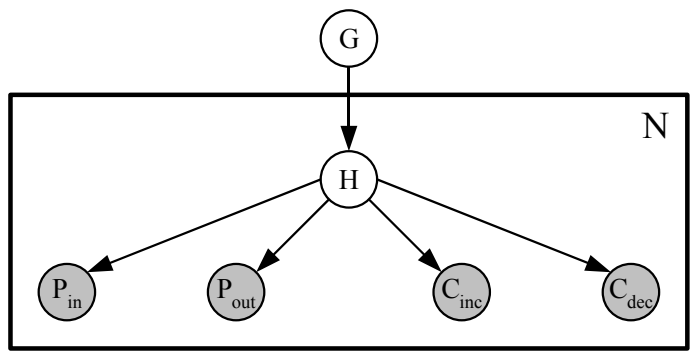

Figure 2: CTBN model for network traffic as a plate model. $N$ is the number of port .

## 4.1 A CTBN Model for Network Traffic

We use the same port-level submodel as our previous work (Xu & Shelton, 2008). We have a latent variable $H$ and four fully observed toggle variables: $P_{in}, P_{out}, C_{inc}, C_{dec}$.

The nodes packet-in, $P_{in}$, and packet-out, $P_{out}$, represent the transmission of a packet to or from the host. They have no intrinsic state: the transmission of a packet is an essentially instantaneous event. Therefore they have events (or "transitions") without having state. This is modeled using a toggle variable in which an event is evidence of a change in the state of the variable and the rate of transition associated with each state is required to be the same.

The nodes connection-increase $C_{in}$ and connection-decrease $C_{dec}$ together describe the status of the number of concurrent connections $C$ active on the host. Notice that $C$ can only increase or decrease by one at any given event (the beginning or ending time of a connection). We assume that the arrival of a new connection and the termination of an existing connection are both independent of the number of other connections. Thus the intensity with which *some* connection starts (or stops) is same as any other connections. Therefore, these are also modeled as toggle variables.

Node $H$ has 8 states that represent different abstract attributes about the machine's internal state. The toggle variables ($P_{in}, P_{out}, C_{inc}$ and $C_{dec}$) are each allowed to change only for 2 of the states of $H$ and they are required to have the same rate for both of these states. 2 hidden states per toggle variable was chosen as a balance between expressive power and model efficiency.

In previous work, we assumed that the traffic associated with different ports are independent of each other, so the port-level submodels are isolated. Here we remove this restriction by introducing another latent variable $G$ that ties the port submodels together. The full model is shown in Figure 2.

## 4.2 Parameter Learning Using RBPF

To calculate the expected sufficient statistics in the E-step of EM for parameter learning, the exact inference algorithm of Nodelman et al. (2002) flattens all the variables into a joint intensity matrix and reasons about the resulting homogeneous Markov process. The time complexity is exponential in the number of variables. For example, if there are 9 port models, the network contains 46 variables in total. Approximate inference techniques like the clique tree algorithm (Nodelman et al., 2002), message passing algorithms (Nodelman et al., 2005; Saria et al., 2007), importance sampling (Fan

& Shelton, 2008) and Gibbs sampling (El-Hay et al., 2008) overcome this problem by sacrificing accuracy.

We notice that our model has a nice tree structure which makes Rao-Blackwellized particle filtering (RBPF) a perfect fit. RBPF uses a particle filter to sample a portion of the variables and analytically integrates out the rest. It decomposes the model structure efficiently and thus reduces the sampling space.

If we denote the $N$ port-level hidden variables as $H_1, ..., H_N$, the posterior distribution of the whole model can be factorized as $P(G, H_1, ..., H_N \mid \tau) = P(G \mid \tau) \prod_{i=1}^{N} P(H_i \mid G, \tau)$. Note that $G$ and $H_i$ are processes, so this probability is a density over complete trajectories. We use a particle filter to estimate $G$'s conditional distribution $P(G \mid \tau)$ as a set of sampled trajectories of $G$. It is difficult to sample directly from the posterior distribution, so we use an importance sampler to sample a particle from a proposal distribution and the particles are weighted by the ratio of its likelihood under the posterior distribution to the likelihood under the proposal distribution (Doucet et al., 2000). Since the variable $G$ is latent and has no parents, we can use forward sampling to sample the particles from $P(G)$ and the weight of each particle is simply the likelihood of $\tau$ conditioned on this trajectory for $G$ (Fan & Shelton, 2008). Each port-level submodel is then $d$-separated from the rest of the network, given full trajectory of $G$ (see Section 3.9 for $d$-separation in CTBNs). Since each is small (only 8 hidden states), they can be marginalized out exactly. That is, we can calculate $P(\tau_i \mid G)$ (where $\tau_i$ is the portion of the trajectory for submodel $i$) exactly, marginalizing out $H_i$ with the $\alpha$-$\beta$ recursions from Section 3.7.

The expected sufficient statistics (ESS) for any variable $X$ in a CTBN are $\bar{T}_{X|\mathbf{U}}[x|\mathbf{u}]$, the expected amount of time $X$ stays at state $x$ given its parent instantiation $\mathbf{u}$, and $\bar{M}_{X|\mathbf{U}}[x, x'|\mathbf{u}]$, the expected number of transitions from state $x$ to $x'$ given $X$'s parent instantiation $\mathbf{u}$. Let $g^i \sim P(G)$, $i = 1, \ldots, M$ be the particles. We define their likelihood weights to be $w_i = \frac{P(g^i|\tau)}{P(g^i)}$ and let $W = \sum_i w_i$ be the sum of the weights. Then general importance sampling allows that an expected sufficient statistic can be estimated in the following way, where $SS$ is any sufficient statistic:

$$
\begin{aligned}
E_{(g,h_1,\ldots,h_N) \sim P(G,H_1,\ldots,H_N|\tau)}&[SS(g, h_1, \ldots, h_N)] \\
&= E_{g \sim P(G|\tau)} E_{h_1,\ldots,h_N \sim P(H_1,\ldots,H_N|g,\tau)}[SS(g, h_1, \ldots, h_N)] \\
&\approx \frac{1}{W} \sum_i w_i E_{h_1,\ldots,h_N \sim P(H_1,\ldots,H_N|g^i,\tau)}[SS(g^i, h_1, \ldots, h_N)] \; .
\end{aligned}
$$

The expected sufficient statistics of the whole model are in two categories: those that depend only on $g$, $ESS(g)$, and those that depend on a port model $k$, $ESS(g, h_k, \tau_k)$. $ESS(g)$ is simply the summation of counts (the amount of time $G$ stays at some state, or the number of times $G$ transitions from one state to another) from the particles, weighted by the particle weights:

$$
E_{g \sim P(G|\tau)}[SS(g)] \approx \frac{1}{W} \sum_i w_i SS(g^i) \; . \tag{6}
$$

---

**Function Wholemodel_Estep**
    **input**: current model $\theta^t$, evidence $\tau$
    **output**: Expected sufficient statistics $ESS$
    $ESS := \{ESS(g), ESS(s_1, g), \ldots, ESS(s_n, g)\}$
    Initialize $ESS$ as empty
    For each particle $g^i \in \{g^1, \ldots, g^M\}, g^i \sim P(G)$
        For each $S_j \in \{S_1, \ldots, S_N\}$
          $[P(\tau_j | g^i), ESS(s_j, g^i)] = $ **Submodel_Estep**$(g^i, \theta^t[S_j], \tau_j)$
        For each $S_j \in \{S_1, \ldots, S_N\}$
          $ESS(s_j, g) = ESS(s_j, g) + \prod_{k \neq j} P(\tau_k | g^i) \times ESS(s_j, g^i)$
        $ESS_{g^i} = $ **CountGSS**$(g^i)$
        $ESS(g) = ESS(g) + \prod_j P(\tau_j | g^i) \times ESS_{g^i}$
    Return $ESS$

---

Figure 3: Rao-Blackwellized particle filtering Estep for the whole model

$ESS(g, h_k, \tau_k)$ can be calculated for each submodel independently:

$$E_{g, h_1, \ldots, h_N \sim P(G, H_1, \ldots, H_N | \tau)}[SS(g, h_k, \tau_k)]$$

$$\approx \frac{1}{W} \sum_i w_i \int_{h_k} P(h_k | g^i, \tau_k) SS(g^i, h_k, \tau_k) \, dh_k$$

$$= \frac{1}{W} \sum_i \frac{\prod_j P(\tau_j | g^i)}{P(\tau)} \int_{h_k} P(h_k | g^i, \tau_k) SS(g^i, h_k, \tau_k) \, dh_k$$

$$\propto \frac{1}{W} \sum_i \prod_{j \neq k} P(\tau_j | g^i) \int_{h_k} P(h_k, \tau_k | g^i) SS(g^i, h_k, \tau_k) \, dh_k \ . \quad (7)$$

The integrals are over all possible trajectories for the hidden process $H_k$. The first line holds by $d$-separation (we need only average over the submodel $k$, given an assignment to $G$). The second line expands the weight. The last line combines the weight term for submodel $k$ with the terms in the integral to get the likelihood of $h_k$ and the submodel data. The constant of proportionality will cancel in the subsequent maximization, or it can be reconstructed by noting that $\sum_x \bar{T}_{X | \mathbf{U}}[x | \mathbf{u}]$ should be the total time of the interval.

This last integral, $\int_{h_k} P(h_k, \tau_k | g^i) SS(g^i, h_k, \tau_k) \, dh_k$, and $P(\tau_j | g^i)$ can be calculated using the technique described by Nodelman et al. (2005), for exact ESS calculation. The calculations are very similar the integrals of Section 3.7, except they intensity matrices can change from interval to interval (they are a function of the sampled trajectory $g_i$).

The full E-step algorithm is shown in Figure 3 ($s_k$ represents all of the variables in submodel $k$). Function **Submodel_Estep** calculates the expected sufficient statistics and the likelihood for a subnet model (Equation 7). Function **CountGSS** counts the empirical time and transition statistics from the sampled trajectory of $G$ (Equation 6).

In EM, we use the ESS as if they were the true sufficient statistics to maximize the likelihood with respect to the parameters. For a "regular" CTBN variable $X$ (such as our hidden variable $G$ and $H$), Equation 5 performs the maximization. For our toggle variables, *e.g.* $P_i$, the likelihood component for the toggle variable is

$$\prod_{\mathbf{u}} Q_{P_i|\mathbf{u}}^{M_{P_i}} \exp(-Q_{P_i|\mathbf{u}} T[U = \mathbf{u}])$$

which can be found by setting $q_{x|\mathbf{u}}$ to be the same value ($Q_{P_i|\mathbf{u}}$) for all $x$ (tieing the parameters) and simplifying the product over $x$ in Equation 2. Thus the maximum likelihood parameter estimate is

$$Q_{P_i|\mathbf{u}} = \frac{M_{P_i}}{T[U = \mathbf{u}]}$$

where $M_{P_i}$ is the number of events for variable $P_i$ and $Q_{P_i|\mathbf{u}}$ is the only parameter: the rate of switching.

We synchronize the particles at the end of each "window" (see Section 6.1) and resample as normal for a particle filter at those points. That is, we propagate the particles forward, but stop them all at the end of the window, resample based on the weights, and then continue with the new set of particles. In general, the particles are not aligned by time, except at these resampling points.

### 4.3 Online Testing Using Likelihood

Once the CTBN model has been fitted to historic data, we detect attacks by computing the likelihood of a window of the data (see Section 6.1) under the model. If the likelihood falls below a threshold, we flag the window as anomalous. Otherwise, we mark it as normal.

In our experiments, we fix the window to be of a fixed time length, $T_w$. Therefore, if the window of interest starts at time $T$, we wish to calculate $p(\tau[T, T + T_w] \mid \tau[0, T])$ where $\tau[s, t]$ represents the observed connections and packets from time $s$ to time $t$. Again, we use a RBPF to estimate this probability. The samples at time $T$ represent the prior distribution $P(G \mid \tau[0, T])$. Propagating them forward across the window of length $T_w$ produces a set of trajectories for $G$, $g^i$. Each submodel $k$ can evalute $P(\tau_k[T, T + T_w] \mid g^i)$ by exact marginalization (the sum of the vector $\alpha_{T+T_w}$, the forward message). The weighted average (over samples $g^k$) of the product of the submodel probabilities is our estimate of $P(\tau[T, T + T_w] \mid \tau[0, T])$.

## 5. Anomaly Detection Using System Calls

Now we turn to the problem of detecting anomalies using system call logs.

### 5.1 A CTBN Model for System Calls

System call logs monitor the kernel activities of machines. They record detailed information of the sequence of system calls to operating system. Many malicious attacks on the host can be revealed directly from the internal logs.

We analyze the audit log format of SUN's Solaris Basic Security Module (BSM) praudit audit logs. Each user-level and kernel event record has at least three tokens: header, subject, and return. An event begins with a "header" in the format of: header, record length in bytes, audit record version number, event description, event description modifier, time and date. The "subject" line consists of: subject, user audit ID, effective user ID, effective group ID, real user ID, real group ID, process ID, session ID, and terminal ID consisting of a device and machine name. A "return" with a return value indicating the success of the event closes the record.

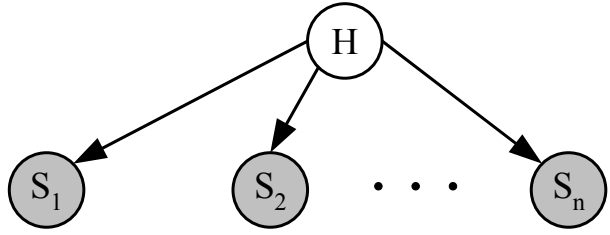

Figure 4: CTBN model for system call data

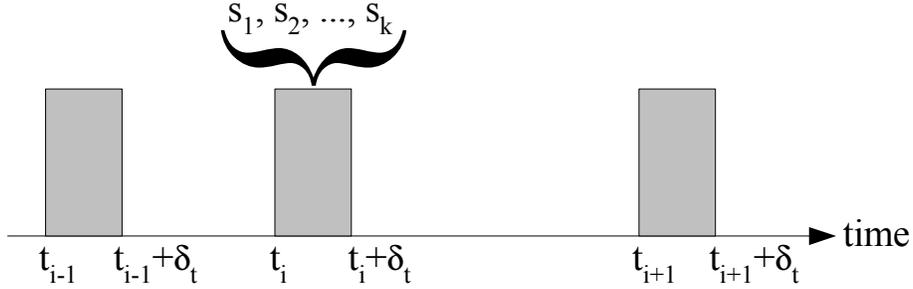

Figure 5: System call traces with a finite resolution clock (resolution = $\delta_t$)

We construct a CTBN model similar to our port-level network model. Individual system calls $S_1, ..., S_N$, which are the event description fields in the header token, are transiently observed: they happen instantaneously with no duration. We treat them as toggle variables like packets in the network model. We also introduce a hidden variable $H$ as a parent of the system calls variables to allow correlations among them. This hidden variable is designed to model the internal state of the machine, although such a semantic meaning is not imposed by our method. Put together, our system call model looks like Figure 4.

If the state space of the hidden variable $H$ is of size $m$, the transition rate matrix of $H$ is

$$\mathbf{Q_H} = \begin{bmatrix} -q_{h_1} & q_{h_1 h_2} & \cdots & q_{h_1 h_m} \\ q_{h_2 h_1} & -q_{h_2} & \cdots & q_{h_2 h_m} \\ \vdots & \vdots & \ddots & \vdots \\ q_{h_m h_1} & q_{h_m h_2} & \cdots & -q_{h_m} \end{bmatrix}.$$

and the transition intensity rate of the toggle variable $s \in S$ given the current value of its parent $H$ is $q_{s|h_i}, i = 1, ..., m$.

To estimate the CTBN model parameters, we again use the expectation maximization (EM) algorithm. The expected sufficient statistics we need to calculate for our model are

- $\bar{M}_{h_i h_j}$, the expected number of times $H$ transitions from state $i$ to $j$;
- $\bar{T}_{h_i}$, the expected amount of time $H$ stays in state $i$; and
- $\bar{M}_{s|h_i}$, the expected number of times system call $s$ is evoked when $H$ is in state $i$.

The maximum likelihood parameters are

$$q_{h_i h_j} = \frac{\bar{M}_{h_i h_j}}{\bar{T}_{h_i}}$$

$$q_{s|h_i} = \frac{\bar{M}_{s|h_i}}{\bar{T}_{h_i}} \; .$$

## 5.2 Parameter Estimation with Finite Resolution Clocks

Because of the finite resolution of computer clocks, multiple instantaneous events (system calls) occur within a single clock tick. Therefore in the audit logs, a batch of system calls may be recorded as being executed at a same time point, rather than their real time stamp, as a result of this finite time accuracy. However, the correct order of the events is kept in the logs. That is, we know exactly that system call $S_2$ follows $S_1$ if they are recorded in this order in the audit logs. Thus all the system call timings are only partially observed. This type of partial observation has not previously been considered in CTBN inference. A typical trajectory $\tau$ over $[0, T]$ of system call data is shown in Figure 5: a batch of system calls are evoked at some time after $t_i$ but before the next clock tick, followed by a quiet period of arbitrary length, and yet another bunch of events at some time after $t_{i+1}$ and so on.

Let $\tau_{t1:t2}$ denote the evidence over interval $[t1, t2)$, $\tau_{t1:t2+}$ denote the evidence over $[t1, t2]$, and $\tau_{t1-:t2}$ denote the evidence over $(t1, t2)$. We define the vectors

$$\alpha_{t_i}^- = p(H_{t_i^-}, \tau_{0:t_i})$$

$$\beta_{t_i}^+ = p(\tau_{t_i^+:T}|H_{t_i^+})$$

where $H_{t_i^-}$ is the value of $H$ just prior to the transition at $t_i$, and $H_{t_i^+}$ is value just afterward. We also define the vectors

$$\alpha_{t_i} = p(H_{t_i}, \tau_{0:t_i^+})$$

$$\beta_{t_i} = p(\tau_{t_i:T}|H_{t_i})$$

where the evidence at the transition time $t_i$ is included. We follow the forward-backward algorithm to compute $\alpha_{t_i}$ and $\beta_{t_i}$ for all $t_i$ at which there is an event. To do this, we split any interval $[t_i, t_{i+1})$ into a "spike" period $[t_i, t_i + \delta_t)$ ($\delta_t$ is one resolution clock), during which there is a batch of system calls, and a "quite" period $[t_i + \delta_t, t_{i+1})$ over which no events exist, and do the propagations separately.

For a "spike" period $[t_i, t_i + \delta_t)$, if the observed event sequence is $s_1, s_2, ..., s_k$, we construct an artificial Markov process $X$ with the following intensity matrix.

$$\mathbf{Q_X} = \begin{bmatrix} \hat{\mathbf{Q}}_\mathbf{H} & \mathbf{Q_1} & \mathbf{0} & \dots & \mathbf{0} \\ \mathbf{0} & \hat{\mathbf{Q}}_\mathbf{H} & \mathbf{Q_2} & \dots & \mathbf{0} \\ \vdots & \vdots & \ddots & \vdots & \vdots \\ \mathbf{0} & \mathbf{0} & \dots & \hat{\mathbf{Q}}_\mathbf{H} & \mathbf{Q_k} \\ \mathbf{0} & \mathbf{0} & \dots & \mathbf{0} & \hat{\mathbf{Q}}_\mathbf{H} \end{bmatrix} \; .$$

where

$$\hat{\mathbf{Q}}_{\mathbf{H}} = \begin{bmatrix} -q_{h_1} - \sum_{s \in S} q_{s|h_1} & q_{h_1 h_2} & \cdots & q_{h_1 h_m} \\ q_{h_2 h_1} & -q_{h_2} - \sum_{s \in S} q_{s|h_2} & \cdots & q_{h_2 h_m} \\ \vdots & \vdots & \ddots & \vdots \\ q_{h_m h_1} & q_{h_m h_2} & \cdots & -q_{h_m} - \sum_{s \in S} q_{s|h_m} \end{bmatrix}$$

and

$$\mathbf{Q_i} = \begin{bmatrix} q_{s_i|h_1} & 0 & \cdots & 0 \\ 0 & q_{s_i|h_2} & \cdots & 0 \\ \vdots & \vdots & \ddots & \vdots \\ 0 & 0 & 0 & q_{s_i|h_m} \end{bmatrix}$$

$X$ tracks the evidence sequence $s_1 \to s_2 \to \ldots \to s_k$. $\mathbf{Q_X}$ is a square block matrix of dimension $m \times (k+1)$. Each block is an $m \times m$ matrix. The subsystem $X$ has $k + 1$ blocks of states. The first block represents the state of $H$ before any events. The second block represents $H$ after exactly one event, $s_1$, happens. The third block represents $H$ after $s_1$ followed by $s_2$ happens, and so on. The last block represents $H$ after all the events finish executing in order. The subsystem has zero transition intensities everywhere except along the sequence pass. The diagonal of $\hat{\mathbf{Q}}_{\mathbf{H}}$ is the same matrix as that of $\mathbf{Q_H}$ except that the transition intensities of all the system call variables are subtracted. This is because the full system includes transitions that were not observed. While those transition rates were set to zero (to force the system to agree with the evidence), such conditioning does not change the diagonal elements of the rate matrix (Nodelman et al., 2002). Within each of the $k + 1$ states of a block, $H$ can freely change its value. Therefore, the non-diagonal elements of $\hat{\mathbf{Q}}_{\mathbf{H}}$ have the same intensities as $\mathbf{Q_H}$. Upon transitioning, $X$ can only transit from some state to another according to the event sequence. Therefore, most of the blocks are $0$ matrices except those to the immediate right of the diagonal blocks. The transition behavior is described by the matrix $Q_i$. $Q_i$ has $0$ intensities on non-diagonal entries because $H$ and $S$ can not change simultaneously. The diagonal element $Q_i(h, h)$ is the intensities of event $s_i$ happening, given the current value of the hidden state is $h$.

We take the forward pass as an example to describe the propagation; the backward pass can be performed similarly. Right before $t_i$, $\alpha_{t_i}^-$ has $m$ dimensions. We expand it to $m(k+1)$ dimensions to form $\alpha_{t_i}$ which only has non-zero probabilities in the first $m$ states. $\alpha_{t_i}$ now describes the distribution over the subsystem $X$. $\alpha_{t_i} e^{\mathbf{Q_X} \delta_t}$ represents the probability distribution at time $t_i + \delta_t$, given that some prefix of the observed sequence occurred. We take only the last $m$ state probabilities to condition on the entire sequence happening, thus resulting in an $m$-dimensional vector, $\alpha_{t_i + \delta_t}$.

For a "quiet" period $[t_i + \delta_t, t_{i+1})$, no evidence is observed. Therefore $\alpha_{t_i + \delta_t}$ is propagated to $\alpha_{t_{i+1}}$ using $\hat{\mathbf{Q}}_{\mathbf{H}}$, the rate matrix conditioned on only $H$ events occuring:

$$\alpha_{t_{i+1}} = \alpha_{t_i + \delta_t} \exp(\hat{\mathbf{Q}}_{\mathbf{H}}(\mathbf{t_{i+1}} - \mathbf{t_i} - \delta_{\mathbf{t}})) \ .$$

When we are done with the full forward-backward pass over the whole trajectory, we can calculate the expected sufficient statistics $\bar{M}_{h_i h_j}$, $\bar{T}_{h_i}$ and $\bar{M}_{s|h_i}$. Again, we refer to the work of Nodelman et al. (2005) for the algorithm.

## 5.3 Testing Using Likelihood

Once we have learned the model from the normal process in the system call logs, we calculate the log-likelihood of a future process under the model. The log-likelihood is then compared to a predefined threshold. If it is below the threshold, a possible anomaly is indicated. With only a single hidden variable, these calculations can be done exactly.

## 6. Evaluation

To evaluate our methodology, we constructed experiments on two different types of data: network traffic traces and system call logs. In the following sections, we show the experiment results on both tasks.

A dynamic Bayesian network (DBN) is another popular technique for graphical modeling of temporal data. Because they slice time, events without state changes (instantaneous events) are difficult to model. Any reasonable time resolution will result with multiple events for the same variable over one time period. There is no standard way of encoding this in a DBN. If we use a toggle variable, it only records the parity of the number of events over the time interval. Furthermore, for the NIDS, events are *very* bursty. During active times, multiple packets are emited per second. During inactive times, there may be no activity for hours. Finding a suitable sampling rate that maintains the efficency of the model is difficult. For the HIDS, the problem is more acute. We do not know of any way of modeling timing ambiguity in a DBN without throwing away all timing information or adding a mathematical framework that essentially turns the DBN into the CTBN described here. In general, we could not find a suitable way to apply a DBN to these problems without essentially turning the DBN into a CTBN by *very* finely slicing time and then applying numeric tricks to speed up inference that amount to converting the stochastic matrices into rate matrices and using numeric integration for the matrix exponential.

We have compared against current adaptive methods for each problem individually. These include nearest neighbor, support vector machines, and sequence time-delaying embedding. We give further details on these methods below.

## 6.1 Experiment Results on Network Traffic

In this section, we present our experiment results on NIDS.

### 6.1.1 DATASETS

We verify our approach on two publicly available real network traffic trace repositories: the MAWI working group backbone traffic MAWI and the LBNL/ICSI internal enterprise traffic LBNL.

The MAWI backbone traffic is part of the WIDE project which has collected raw daily packet header traces since 2001. It records the network traffic through the inter-Pacific tunnel between Japan and the USA. The dataset uses `tcpdump` and IP anonymizing tools to record 15-minute traces every day, and consists mostly of traffic from or to Japanese universities. In our experiment, we use the traces from January 1st to 4th of 2008, with 36,592,148 connections over a total time of one hour.

The LBNL traces are recorded from a medium-sized site, with emphasis on characterizing internal enterprise traffic. Publicly released in an anonymized form, the LBNL data collects more than

# packets flowing from source to destination
# packets flowing from destination to source

---

# connections by the same source in the last 5 seconds
# connections to the same destination in the last 5 seconds
# different services from the same source in the last 5 seconds
# different services to the same destination in the last 5 seconds

---

# connections by the same source in the last 100 connections
# connections to the same destination in the last 100 connections
# connections with the same port and source in the last 100 connections
# connections with the same port and destination in the last 100 connections

Figure 6: Features for nearest neighbor approach from the work of (Lazarevic et al., 2003).

100 hours network traces from thousands of internal hosts. From what is publicly released, we take one hour traces from January 7th, 2005 (the latest date available), with 3,665,018 total connections.

### 6.1.2 WORM DETECTION

We start with the problem of worm detection. We split traffic traces for each host: half for training and half for testing. We learn a CTBN model from the training data for each of the hosts. Since the network data available are clean traffic with no known intrusions, we inject real attack traces into the testing data. In particular, we inject IP Scanner, W32.Mydoom, and Slammer. We then slide a fixed-time window over the testing traces, report a single log-likelihood value for each sliding window, and compare it with a predefined threshold. If it is below the threshold, we predict it as an abnormal time period. We define the ground truth for a window to be abnormal if any attack traffic exists in the interval, and normal otherwise. The window size we use is 50 seconds. We only consider windows that contain at least one network event.

We compare our method employing RBPF with our previous factored CTBN model (Xu & Shelton, 2008), connection counting, nearest neighbor, Parzen-window detector (Yeung & Chow, 2002), and one-class SVM with a spectrum string kernel (Leslie, Eskin, & Noble, 2002).

The connection counting method is straightforward. We score a window by the number of initiated connections in the window. As most worms aggregate many connections in a short time, this method captures this particular anomaly well.

To make nearest neighbor competitive, we try to extract a reasonable set of features. We follow the feature selection of the work of Lazarevic et al. (2003), who use a total of 23 features. Not all of their features are available in our data. Those available are shown in Figure 6. Notice that these features are associated with each connection record. To apply the nearest neighbor method to our window based testing framework, we first calculate the nearest distance of each connection inside the window to the training set (which is composed of normal traffic only), and assign the maximum among them as the score for the window. Similarly, for the Parzen window approach, we apply the same feature set and assign the maximum density among all the connections inside a window to be the score of that window.

Besides the above feature-based algorithms, we would also like to see how sequence-based approaches compare against our methods. These algorithms are widely used in network anomaly detection. Like our approach, they treat the traffic traces as stream data so that sequential contexts can be explored. One-class SVM with spectrum string kernel was chosen for comparison. We implemented a spectrum kernel in the LIBSVM library (Chang & Lin, 2001). We give the network activities (such as a connection starting or ending, or a packet emmision or receipt) inside each port-level submodel a distinct symbol. The sequence of these symbols are fed to the algorithm as inputs. A decision surface is trained from normal training traffic. In testing, for each sliding window, the distance from this window string to the decision hyperplane is reported as the window score. We also tried experiments using the edit distance kernel, but their results are dominated by the spectrum kernel, so we do not report them here.

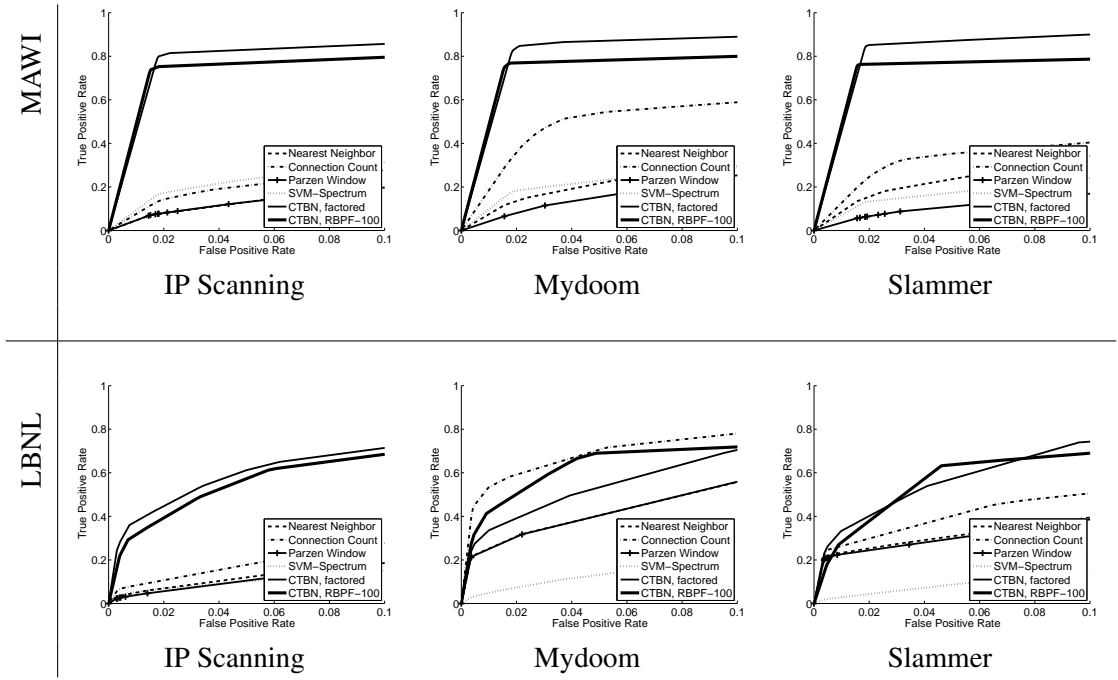

Figure 7: ROC curves of testing results on IP scanning attack, Mydoom attack and Slammer attack. $\beta = 0.001$. Top: MAWI. Bottom: LBNL.

When injecting the attack traffic, we randomly pick a starting point somewhere in the first half of the test trace and insert worm traffic for a duration equal to $\alpha$ times the length of the full testing trace. The shorter $\alpha$ is, the harder it is to detect the anomaly. We choose $\alpha$ to be 0.02% for all the experiments in this work to challenge the detection tasks. We also scaled back the rates of the worms. When running at full speed, a worm is easy to detect for any method. When it slows down (and thus blends into the background traffic better), it becomes more difficult to detect. We let $\beta$ be the scaling rate (e.g. 0.1 indicates a worm running at one-tenth of its normal speed).

For our method, we set the state space of variable $G$ to be 4 and variable $H$ to be 8. We use 100 samples for particle filtering, and resample the particles after every 50 seconds. For the SVM spectrum kernel method, we choose the sub-sequence length to be 5 and the parameter $\nu$ to be 0.8.

We show the ROC curves of all the methods in Figure 7. The curves show the overall performance on the 10 most active hosts for each dataset. Each point on the curves corresponds to a different threshold of the algorithm. Our CTBN method out-performs the other algorithms except in the single case of the Mydoom attack against a background of the LBNL traffic. In many cases, the advantages of the CTBN approach are pronounced.

For all of the MAWI data, the factored and non-factored CTBN models perform comparably. We believe this is because the data only captures connections that traverse a trans-Pacific link. Therefore, not all of the connections in or out of a machine are represented. This makes reasoning about the global pattern of interaction for a machine difficult. For the LBNL data, one attack (IP scanning) shows no advantage to a non-factored model. One attack (Mydoom) shows a distinct advantage. And one attack (Slammer) indicates some advantage, depending on the desired false positive rate. This demonstrate some advantage to jointly modeling the traffic across all ports, although it is clear this advantage is not uniform over traffic patterns and attack types.

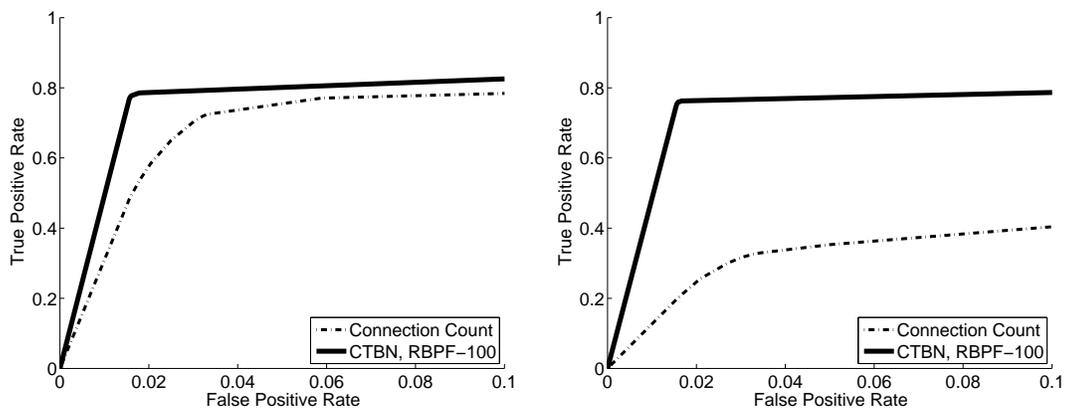

Figure 8: ROC curves of testing results for Slammer attack on MAWI dataset demonstrating the effect of slowing the attack rate. Left: $\beta = 0.01$. Right: $\beta = 0.001$

We also show how the ROC curves shift as we scale back the worm running speed $\beta$ in Figure 8. As firewalls are built to be more sensitive to block malicious traffic, worms have to act more stealthy to sneak through. We demonstrate the robustness of our method compared to the best competitor (connection counts) to the speed of the worm's attack.

### 6.1.3 HOST IDENTIFICATION

Identifying individual hosts based on their network traffic patterns is another useful application of our model. For instance, a household usually installs a network router. Each family member's computer is connected to this router. To the outside Internet, the network traffic going out of the router behaves as if it is coming from one peer, but it is actually coming from different people. Dad will possibly read sports news while kids surf on social networks. It is interesting as well as useful to tell which family member is contributing the current network traffic. Host identification can also be used to combat identity theft. When a network identity is abused by the attacker, host identification techniques can help the network administrator tell whether the current network traffic of this host is consistent with its usual pattern or not.

The first set of experiments we construct is a host model fitting competition. The same 10 hosts picked for the worm detection tasks from LBNL dataset compose our testing pool. We learn the coupled CTBN model for each host. We split the test traces (clean) of a particular host into segments with lengths of 15 seconds. For each of the segments, we compute the log-likelihood of the segment under the learned model from all the hosts (including its own), and label the segment with the host that achieves the highest value. We compute a confusion matrix $C$ whose element $C_{ij}$ equals the fraction of test traces of host $i$ for which model $j$ has highest log-likelihood. We expect to see the highest hit rates fall on the diagonals because ideally a host should be best described by its own model. Table 9 shows our results on the dataset of LBNL. The vast majority of traffic windows are assigned to the correct host. With the exception of host 1, the diagonals are distinctly higher than other elements in the same row. For comparison, we performed the same experiment using SVM spectrum kernel method. Again, we selected the sub-sequence length to be 5 and the parameter $\nu$ to be 0.8. We tried multiple methods for normalization (of the distance to the hyperplane) and variations of parameters. All produced very poor results with almost all of the windows assigned to a single host. We omit the table of results.

| Host | 1 | 2 | 3 | 4 | 5 | 6 | 7 | 8 | 9 | 10 |
|------|------|------|------|------|------|------|------|------|------|------|
| **1** | 0.09 | 0.41 | **0.47** | 0 | 0 | 0 | 0 | 0.03 | 0 | 0 |
| **2** | 0.06 | **0.50** | 0.31 | 0 | 0 | 0.13 | 0 | 0 | 0 | 0 |
| **3** | 0 | 0 | **1** | 0 | 0 | 0 | 0 | 0 | 0 | 0 |
| **4** | 0 | 0 | 0 | **1** | 0 | 0 | 0 | 0 | 0 | 0 |
| **5** | 0 | 0.08 | 0.13 | 0 | **0.68** | 0.06 | 0.03 | 0.01 | 0 | 0.01 |
| **6** | 0 | 0.20 | 0.03 | 0 | 0.01 | **0.74** | 0 | 0.02 | 0 | 0 |
| **7** | 0.25 | 0 | 0.06 | 0.02 | 0 | 0 | **0.66** | 0 | 0 | 0.01 |
| **8** | 0 | 0.08 | 0.28 | 0 | 0 | 0.05 | 0 | **0.59** | 0 | 0 |
| **9** | 0.04 | 0.05 | 0.13 | 0.01 | 0.01 | 0.01 | 0.02 | 0 | **0.73** | 0 |
| **10** | 0 | 0.03 | 0.18 | 0.01 | 0 | 0.15 | 0.03 | 0.03 | 0 | **0.57** |

Figure 9: Confusion matrix for LBNL for host identification using CTBN

Our second experiment is a host traffic differentiation task. We mingle the network traffic from another host with the analyzed host. We expect the detection method to successfully tell apart the two. To verify this idea, we pick one host among the 10 we choose above from LBNL dataset and split its traffic evenly into training and testing. We again learn the model from training data. For testing data, we randomly choose a period and inject another host's traffic as if it were a worm. Our goal is to identify the period as abnormal since the host's traffic is no longer its own behavior. Figure 10 displays the results from two such combination tests. The parameters for injecting the traffic as a worm are $\alpha = 0.02$, $\beta = 0.001$. In the left graph, the nearest neighbor and Parzen window curve overlap, and both CTBN curves overlap. In the right graph, the coupled CTBN curve substantially outperforms all the other curves.

## 6.2 Experiment Results on System Call Logs

In this section, we present our experiment results on HIDS.

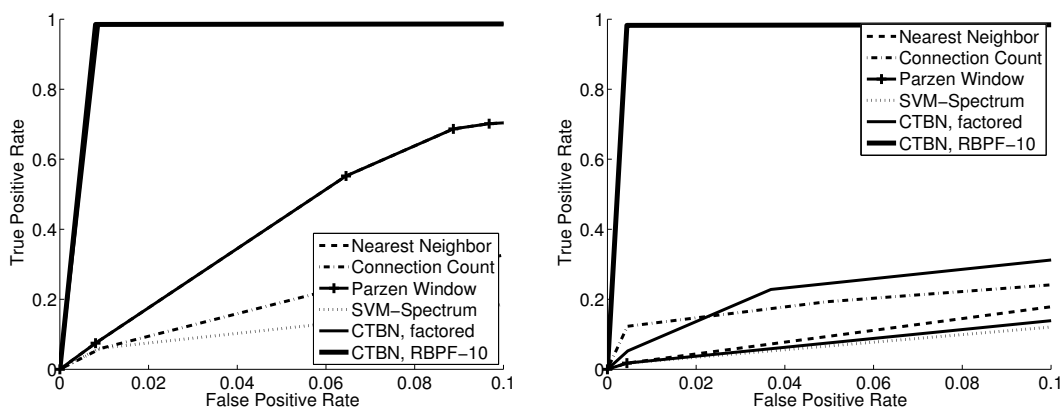

Figure 10: ROC curves of testing results on host identification on the LBNL data. Left: host 1, Nearest neighbor curve and Parzen window curve overlap, both CTBN curves overlap. Right: host 2.

| Week | # normal processes | # attack processes |
|------|--------|--------|
| 1 | 786 | 2 |
| 2 | 645 | 4 |
| 3 | 775 | 20 |
| 4 | 615 | 331 |
| 5 | 795 | 10 |
| 6 | 769 | 24 |
| 7 | 584 | 0 |

| System Call | # occurrence | System Call | # occurrence |
|-------------|--------------|-------------|--------------|
| close | 123403 | execve | 1741 |
| ioctl | 68849 | chdir | 1526 |
| mmap | 60886 | chroot | 328 |
| open | 42479 | unlink | 26 |
| fcntl | 7416 | chown | 23 |
| stat | 6429 | mkdir | 4 |
| access | 2791 | chmod | 1 |

Figure 11: Left: DARPA BSM process summary. Right: DARPA BSM system call summary

### 6.2.1 Dataset

The dataset we used is the 1998 DARPA Intrusion Detection Evaluation Data Set from MIT Lincoln Laboratory. Seven weeks of training data that contain labeled network-based attacks in the midst of normal background data are publicly available at the DARPA website. The Solaris Basic Security Module (BSM) praudit audit data on system call logs are provided for research analysis. We follow Kang, Fuller, and Honavar (2005) to cross-index the BSM logs and produce a labeled list file that labels individual processes. The resulting statistics are shown on the left table of Figure 11. The frequency of all the system calls appearing in the dataset is summarized in descending order on the right of Figure 11.

### 6.2.2 Anomaly Detection

Our experimental goal is to detect anomalous processes. We train our CTBN model on normal processes only and test on a mixture of both normal and attack processes. The state space of the hidden variable $H$ is set to 2. The log-likelihood of a whole process under the learned model represents the score of this process. We compare to the score with a predefined threshold to classify the process as a normal one or a system abuse.

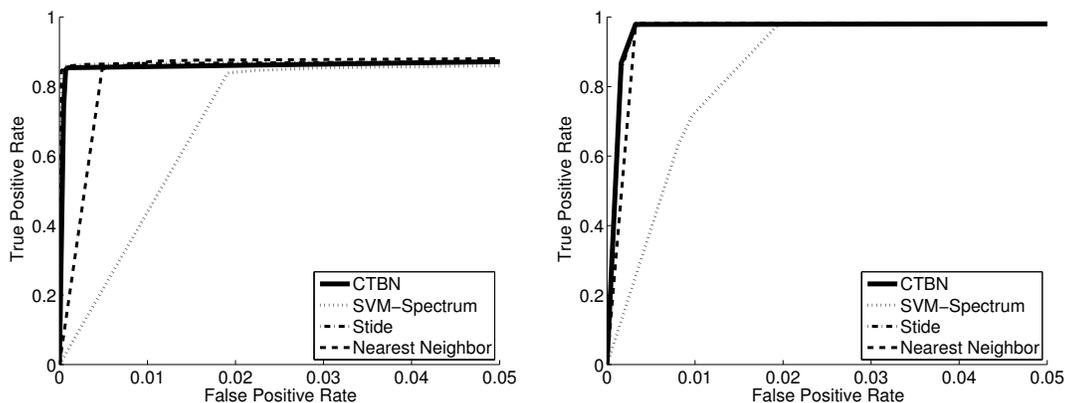

Figure 12: ROC curves for BSM data detection. Left: Training on week 1 and combined testing results on week 2 to 7; Right: Training on week 3 and test on week 4, Stide curve and CTBN curve overlap

We implement sequence time-delaying embedding (stide) and stide with frequency threshold (t-stide) for comparison (Warrender, Forrest, & Pearlmutter, 1999). These two algorithms build a database of all previously seen normal sequences of system calls and compare the testing sequences with it. They are straightforward and perform very well empirically on most of the system call log datasets. We choose the parameter $k$, the sequence length to be 5, and $h$, the locality frame length, to be 50. The results for t-stide are not shown in the following resulting graphs since they overlapped with stide in almost all cases.

Other approaches we compare against are nearest neighbor and one-class SVM with spectrum string kernel and edit distance kernel. We follow Hu et al. (2003) and transform a process into a feature vector, consisting of the occurrence numbers of each system call in the process. The nearest distance between a testing process and the training set of processes is assigned as the score. For one-class SVM, processes are composed of strings of system calls. Normal processes are used for learning the bounding surface and the signed distance to it is assigned as the score. We set the subsequence length to be 5 and the parameter $\nu$ to be 0.5. Again, since the edit distance kernel results are dominated by the spectrum kernel, we do not show them.

Figure 12 displays the results from two experiment settings. In the left graph, we train the model on the normal processes from week 1 and test it on all the processes from weeks 2 to 7. In the right graph, we train on normal processes from week 3 and test it on all the processes from week 4, the richest in attack processes volume. Because attacks are relatively rare compared to normal traffic, we are most interested in the region of the ROC curves with small false positive rates. So we only show the curves in the area where the false positive rate falls in the region [0, 0.05]. Our CTBN method beats nearest neighbor and SVM with spectrum kernel in both experiments. stide performs slightly better than our method in the combined test, but achieves the same accuracy in the experiment using only week 3 and testing on week 4. The advantage to the CTBN model over stide is that it can be easily combined with other prior knowledge and other data sources (such as the network data from NIDS). We demonstrate that there is no loss of performance from such flexibility.

## 7. Conclusions

In the realm of temporal reasoning, we have introduced two additions to the CTBN literature. First, we demonstrated a Rao-Blackwellized particle filter with continuous evidence. Second, we demonstrated that we can learn and reason about data that contains imprecise timings, while still refraining from discretizing time.

In the realm of intrusion detection, we have demonstrated a framework that performs well on two related tasks with very different data types. By concentrating purely on event timing, without the consideration of complex features, we were able to out-perform existing methods. The continuous-time nature of our model aided greatly in modeling the bursty event sequences that occur in systems logs and network traffic. We did not have to resort to time slicing, either producing rapid slices that are inefficient for quite periods, or lengthy slices that miss the timing of bursty events.

A combination of the two sources of information (system calls and network events) would be straight-forward with the model we have produced. We believe it would result in more accurate detection. The collection of such data is difficult, however; we leave it as an interesting next step.

## Acknowledgments

This project was supported by Intel Research and UC MICRO, by the Air Force Office of Scientific Research (FA9550-07-1-0076), and by the Defense Advanced Research Project Agency (HR0011-09-1-0030).